\documentclass[runningheads]{llncs}
\usepackage{amssymb,amsfonts,bm}
\usepackage[colorlinks, linkcolor=magenta, anchorcolor=green, citecolor=blue, urlcolor=black]{hyperref}
\usepackage{array}
\usepackage{textcomp}
\usepackage{stfloats}
\usepackage{url}
\usepackage{verbatim}
\usepackage{xcolor}
\usepackage{amsmath}

\usepackage{bigstrut,multirow,rotating}
\usepackage{booktabs}
\usepackage{pifont}
\usepackage{colortbl}
\usepackage{graphicx}
\usepackage[normalem]{ulem}
\usepackage[T1]{fontenc}
\usepackage{geometry}
\newcommand{\equalcontrib}{\textsuperscript{\textdagger}}

\begin{document}
\title{EcoVLA: Energy-Efficient Device-Edge Co-Inference for Vision-Language-Action Models under Real-Time Constraints}
\titlerunning{EcoVLA: Energy-Efficient Device-Edge Co-Inference for VLA Models}
\author{
Ao Zhou\inst{1}\equalcontrib \and
Bo Dai\inst{1}\equalcontrib \and
Le Yu\inst{1} \and
Xingyu Liu\inst{2} \and
Zeyu Hao\inst{1} \and
Lingkun Long\inst{1} \and \\
Chunming Hu\inst{1} \and
Jianlei Yang\inst{1}
}

\authorrunning{A. Zhou et al.}

\institute{
Beihang University, No.37 Xueyuan Road, Beijing, China\\
\email{aozhou@buaa.edu.cn} \& \email{jianlei@buaa.edu.cn}\\
\and
Beijing University of Technology, No.100 Pingleyuan Road, Beijing, China\\
\equalcontrib Ao Zhou and Bo Dai contributed equally to this work.
}

%
\maketitle              
\begin{abstract}
Vision-Language-Action (VLA) models have emerged as a promising foundation for Embodied AI, but their high inference cost poses significant challenges for deployment in robotic systems. In practice, on-device inference is constrained by limited compute capacity and energy budgets, struggling to simultaneously satisfy real-time control and energy efficiency requirements. Alternatively, offloading the inference workload to an edge server is susceptible to fluctuations in system conditions, introducing unpredictable latency risks. Device-edge co-inference offers a promising solution, but systematic research tailored to VLA models remains scarce, particularly a unified co-inference framework that jointly addresses real-time constraints and system-level energy efficiency. Thus, we propose EcoVLA, an adaptive device-edge co-inference framework for VLA models that maximizes system energy efficiency under real-time constraints. EcoVLA first introduces a unified stage-level abstraction over different VLA paradigms, establishing an architecture-agnostic co-inference design space. It then formulates a joint device-edge-network latency and energy prediction model to enable rapid runtime evaluation of candidate co-inference schemes. Building on this, EcoVLA continuously selects the energy-optimal scheme satisfying real-time constraints with millisecond-level overhead, adapting to runtime variations in network and system states. Furthermore, EcoVLA incorporates a lightweight transmission mechanism for inter-stage intermediate tensors to reduce the communication overhead incurred by cross-device collaboration. 
Experimental results across VLA models show that EcoVLA improves system energy efficiency by up to $236\%$ over existing co-inference approaches under a 20 Hz action output frequency constraint, while consistently maintaining SLO satisfaction under dynamic network and edge workload conditions.

\end{abstract}

\section{Introduction}\label{sec:introduction}

Vision–Language–Action (VLA) models have recently demonstrated impressive capabilities across various embodied AI tasks by integrating visual perception and language understanding into the action generation process~\cite{sapkota2025vision}.
However, the prohibitive computational cost of VLA models is fundamentally mismatched with the limited compute capability and energy budget of robot devices~\cite{wen2025tinyvla}.
A direct example is that deploying the popular OpenVLA model on a Jetson AGX Orin results in per-inference latency of more than 1 second, whereas dynamic interaction scenarios typically require at least 10 Hz to maintain control stability. 
This gap between inference cost and real-time requirements severely limits the practical deployment of VLA models in real-world robotic systems.

Research efforts have been made to address the inefficiency of VLA models on edge devices.
ActionFlow~\cite{dai2025actionflow} introduces a cross-request pipelining strategy that batches the decode and prefill stages across time steps to improve hardware utilization.
RealtimeVLA~\cite{ma2025running} further improves inference performance from a system perspective through techniques such as CUDA Graph and operator-level optimization.
While these methods yield meaningful efficiency gains, they cannot overcome the fundamental resource ceiling of on-device hardware, leaving energy consumption and real-time latency constraints largely unresolved.
As a result, many VLA works instead offload the full model to an edge server, reducing the robot to a thin client responsible only for sensing and action execution~\cite{black2024pi_0,kim2024openvla}. 
However, this naive offloading strategy offers no latency guarantees under network fluctuations and concurrent access from multiple robots, making inference timeouts difficult to avoid.

As an emerging computing paradigm, device-edge co-inference enables collaborative execution of a model across the robot device and the edge server, exposing a richer design space for trading off computation, communication, latency, and energy efficiency~\cite{zhou2024graph}.
It has been widely used in DNN deployment to overcome on-device resource limitations and enable runtime adaptation to dynamic environment, but a dedicated design for VLA models is still missing.
Existing co-inference studies, especially those designed for LLMs~\cite{yuan2025task,qu2025mobile,yang2024efficient}, are not directly applicable because they overlook the characteristics of VLA models and embodied control workloads, and mostly target throughput optimization instead of strict real-time constraints.

More importantly, a key insight motivating our work is that, in robotic control loops, inference speed is not the only metric that matters. Even if inference completes early, the system must wait for the current action execution period before issuing the next command, a structural idle time also identified by VLASH~\cite{tang2025vlash}. 
As a result, system-level energy efficiency should be evaluated over the entire control cycle rather than inferred from instantaneous power or raw inference latency alone. This shifts our core optimization objective: rather than maximizing throughput, maximizing energy efficiency (output actions per joule) under SLO constraints (output actions per second) becomes the metric that determines task endurance and deployment feasibility.
Achieving this objective is non-trivial for several reasons. 
First, VLA paradigms are still evolving, and mainstream autoregressive and diffusion-based models differ significantly in their execution processes and runtime characteristics. 
Second, robotic environments are highly dynamic, under which static co-inference strategies can hardly sustain an effective balance between computation and communication.
Finally, system-level energy efficiency in device-edge systems is affected by multiple coupled factors, making intuition-based estimation likely to yield suboptimal solutions.

To address the above challenges and enable efficient device-edge deployment of VLA models, we propose EcoVLA, the first adaptive device-edge co-inference framework designed for VLA models.
EcoVLA targets a new optimization objective: maximizing Actions/J under real-time constraints.
To achieve this goal, EcoVLA first introduces a unified co-inference design space by decoupling VLA computation graphs into a stage-level execution graph, which provides an architecture-agnostic abstraction together with unified execution and communication support for heterogeneous VLA paradigms.
It then develops a joint latency-energy modeling method for VLA co-inference systems, which captures per-stage latency, power, and cross-device transmission cost on heterogeneous platforms, enabling fast runtime evaluation of end-to-end latency and energy efficiency. Based on this model, EcoVLA further performs energy-priority dynamic scheduling under real-time constraints, continuously selecting the energy-optimal co-inference scheme within the SLO-feasible region with only millisecond-level overhead under changing network and load conditions.
In addition, EcoVLA incorporates a lightweight transmission mechanism for intermediate tensors to further reduce the communication overhead of cross-device execution.

The main contributions of this paper are summarized as follows:
\begin{itemize}
    \item \textbf{Framework.} To the best of our knowledge, EcoVLA is the first paradigm-agnostic adaptive device-edge co-inference framework for VLA models that targets energy efficiency maximization under real-time constraints, supporting unified design and deployment across heterogeneous VLA models and systems.
    \item \textbf{System-Aware Scheduling.} We develop a system performance modeling method that enables accurate and efficient runtime prediction, achieving over $95\%$ prediction accuracy within a $20\%$ error bound and driving dynamic scheduling decisions that continuously select the energy-optimal co-inference configuration within the SLO-feasible design space.
    \item \textbf{Evaluation.} Extensive experiments across diverse VLA models and system configurations show that EcoVLA improves system energy efficiency by up to $236\%$ over fixed co-inference baselines under a $20$ Hz action output frequency constraint, while consistently satisfying the SLO. 
\end{itemize}
\section{Related Work}

\subsection{VLA Deployment and Acceleration} \label{sec:vla_accel}

VLA models mainly follow two design paradigms~\cite{black2024pi_0,kim2024openvla,zhang2025pure_vla_survey}, namely autoregressive and diffusion-based paradigms. Although these paradigms differ substantially in execution flow and computational hotspots, both suffer from a common challenge: their inference cost far exceeds the compute capacity and power budget of on-device platforms, making it difficult to satisfy the real-time requirements of robot control. To improve on-device inference efficiency, existing efforts have evolved from lightweight model design~\cite{wen2025tinyvla,team2024octo} to inference-process and system-level optimizations. For autoregressive VLAs, prior work reduces redundant computation through caching, scheduling, and token pruning~\cite{xu2025vlacache,li2026spvla}; for diffusion-based VLAs, EfficientVLA performs training-free joint compression over the language module, visual tokens, and action head~\cite{yang2025efficientvla}. At the system level, ActionFlow~\cite{dai2025actionflow} and Real-time VLA~\cite{ma2025running} lower on-device latency through operator fusion and pipelined scheduling, while VLASH~\cite{tang2025vlash} overlaps inference with action execution to mitigate structural idle time.
However, existing VLA acceleration methods primarily target inference latency, reaction speed, or control frequency, without explicitly modeling system energy efficiency under continuous robot control scenarios. Moreover, these approaches remain fundamentally device-local optimizations whose gains are still constrained by the compute, memory, and power budgets of on-device platforms.

\subsection{Device-Edge Co-Inference}
\label{sec:co_inference}

Device-edge co-inference partitions inference computation between on-device processors and edge servers to overcome single-device resource limitations. Prior work has progressed from layer-level partitioning for classical DNNs~\cite{kang2017neurosurgeon} to co-design for more complex architectures such as GNNs, where communication-aware mapping and system-level performance prediction are jointly considered~\cite{zhou2024graph,zhou2025acegnn}. More recently, co-inference research has expanded to generative models, with LLM-oriented mechanisms including model sharding~\cite{zhang2024edgeshard}, task-oriented feature compression and operator-level scheduling~\cite{yuan2025task,yang2024efficient}, and distributed speculative verification~\cite{ning2025dssd}.
However, these studies predominantly target applications such as question answering, where response latency tolerance is relatively high, and primarily optimize latency and throughput rather than the joint trade-off between real-time performance and system energy efficiency in continuous robot control scenarios. For robotic applications, sustaining online continuous control and long-duration task execution requires the system to not only meet strict real-time responsiveness but also maintain high energy efficiency. Meanwhile, existing co-inference design spaces are largely built around DNN layer structures or LLM token-generation pipelines, lacking a unified co-inference abstraction and stage-level scheduling support for heterogeneous VLA paradigms such as autoregressive and diffusion-based models.

\section{Motivation and Overview}

\subsection{Why VLA Co-Inference is Challenging}

Unlike conventional model partitioning, building an efficient device-edge co-inference system for VLA models requires addressing three fundamental challenges, which directly motivate the core designs of EcoVLA.

\begin{figure}[t]
    \centering
    \includegraphics[width=1\linewidth]{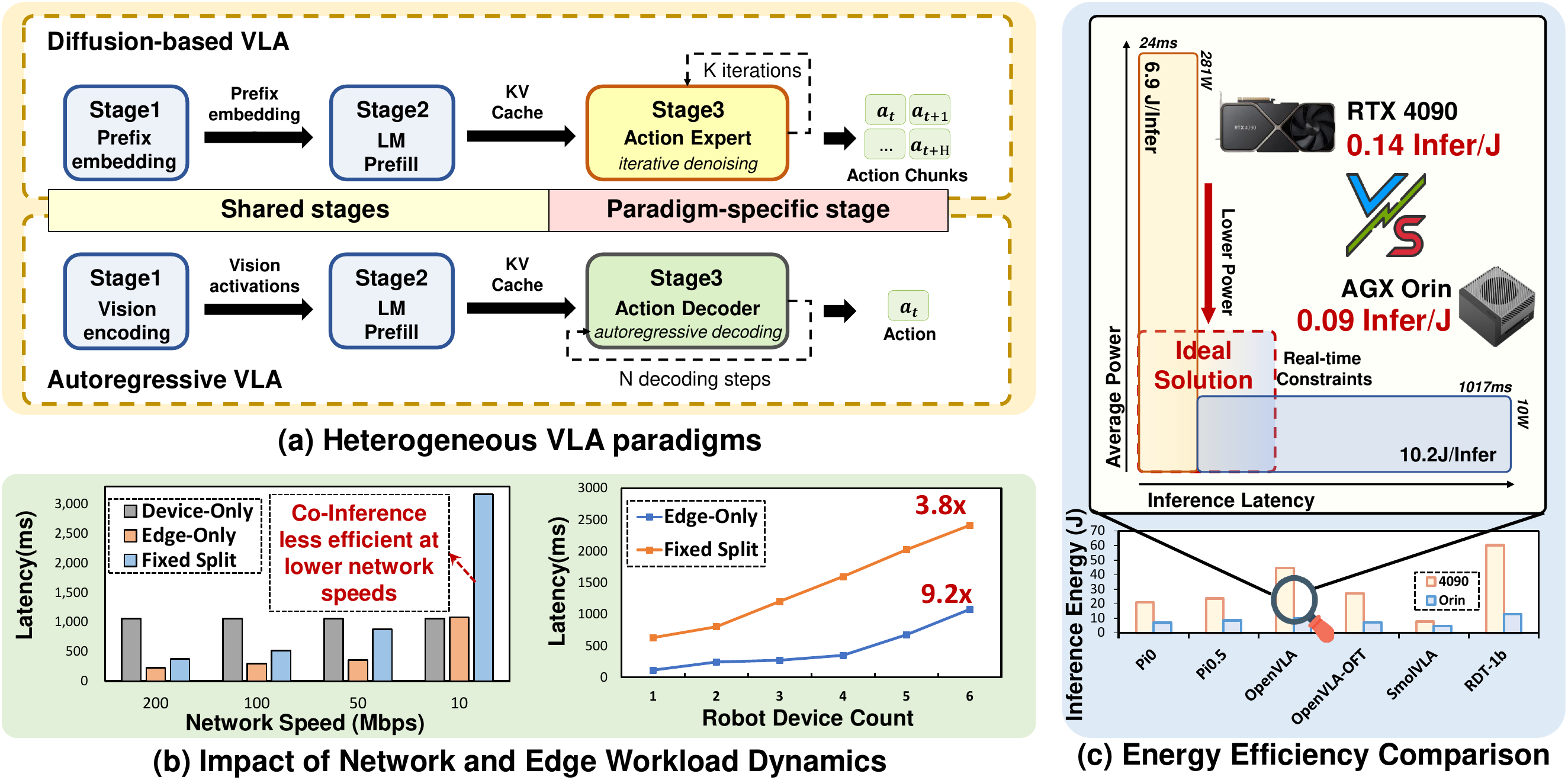}
    \caption{Challenges and Motivation of EcoVLA.}
    \label{fig:motivation}
    \vspace{-6pt}
\end{figure}

\textbf{(1) Heterogeneous VLA paradigms lack a unified co-inference design space and runtime support.}
VLA models are evolving rapidly, and mainstream paradigms differ substantially in both architecture and execution flow, as shown in Fig.~\ref{fig:motivation}(a). Autoregressive VLAs, such as OpenVLA, perform visual encoding, language-conditioned fusion, and step-wise action decoding, whereas diffusion-based VLAs, such as $\pi_0$, generate continuous action trajectories through multi-step iterative denoising. 
These paradigms differ in stage boundaries, intermediate tensor forms, and feasible partition granularity, making a single partitioning strategy difficult to generalize.
Moreover, without a unified runtime abstraction, each new VLA paradigm would further require re-implementing its execution, communication, and scheduling logic from scratch. 
Therefore, the key challenge is not only where to split, but how to build an architecture-agnostic co-inference design space with general and efficient runtime support.

\textbf{(2) Dynamic runtime conditions make static deployment strategies unreliable.}
Unlike typical LLM serving scenarios, VLA deployment operates under highly dynamic conditions, where network fluctuations and edge workload variations continuously reshape system behavior.
As shown in Fig.~\ref{fig:motivation}(b), for OpenVLA inference, neither pure edge offloading nor fixed device-edge partitioning~\cite{jiang2026fast} can reliably satisfy real-time requirements over time.
For Edge-Only inference, degraded networks or increased edge contention can quickly inflate tail latency and cause timeouts. 
For fixed partitioning, a once-effective split can become suboptimal or even infeasible as the balance between computation and communication shifts. 
In essence, VLA co-inference requires continuously balancing computation and communication costs, which static strategies cannot sustain. 
This calls for system-aware online adaptation to runtime changes.

\textbf{(3) System energy efficiency cannot be inferred from device power or inference latency alone, making intuition-driven partitioning unreliable.}
For VLA deployment, the proper objective is system-level energy efficiency under real-time constraints.
A common intuition is to place compute-intensive stages on lower-power devices, assuming that lower power implies higher energy efficiency.
However, this assumption is often misleading.
For example, although Jetson AGX Orin consumes far less power than an RTX 4090 GPU, the latter can deliver nearly an order-of-magnitude lower inference latency. 
As a result, the higher-power GPU may still achieve better energy efficiency during VLA inference, as shown in Fig.~\ref{fig:motivation}(c). 
This example shows that computing fast and computing efficiently are nonlinearly related, and the energy-optimal configuration depends on the joint interaction of device capability, edge compute, network overhead, and real-time constraints, shifting with runtime conditions. Identifying such a scheme thus requires a unified model that jointly captures per-stage latency, power, and communication cost, and supports millisecond-level evaluation within the SLO-feasible region to drive runtime scheduling.

\begin{figure}[t]
    \centering
    \includegraphics[width=1\linewidth]{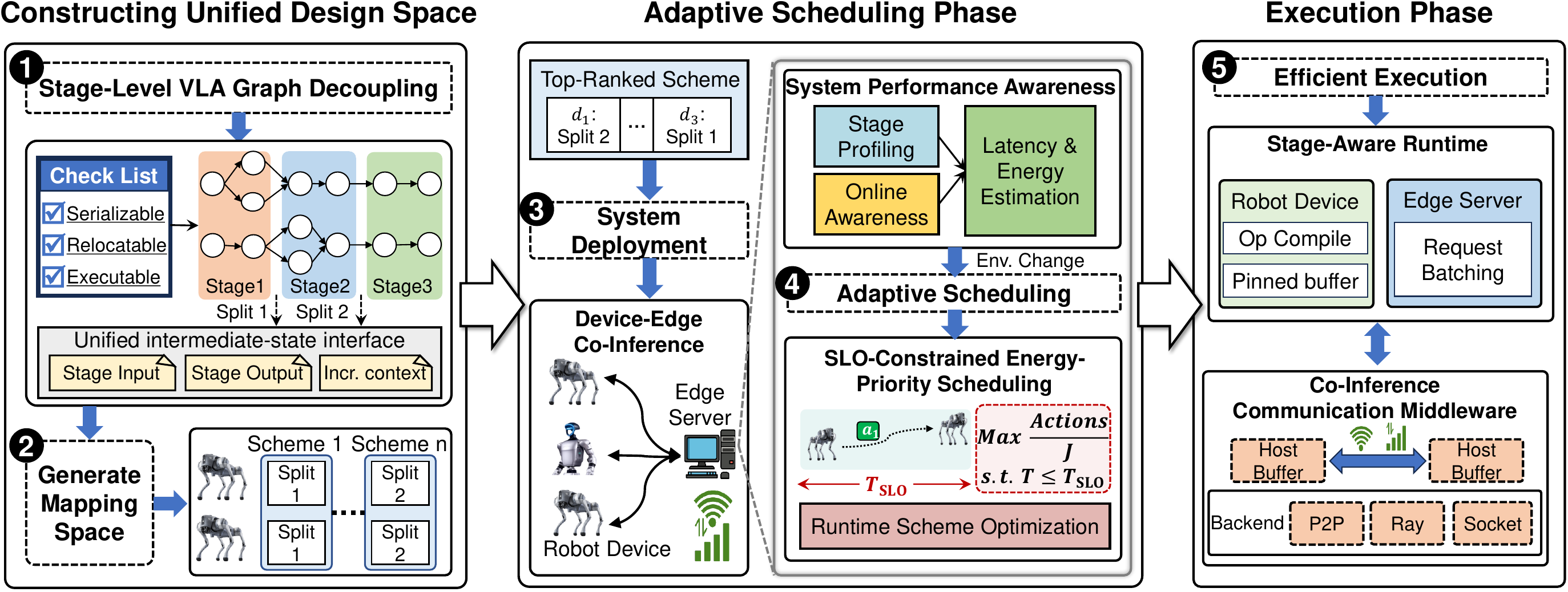}
    \caption{Overview of EcoVLA framework.}
    \label{fig:framework}
    \vspace{-6pt}
\end{figure}

\subsection{Why EcoVLA Achieves Energy-Efficient Real-Time Co-Inference}

Fig.~\ref{fig:framework} shows an overview of the proposed EcoVLA framework, which addresses the above challenges in a unified manner.
EcoVLA first constructs a unified stage-level co-inference design space that decouples heterogeneous VLA paradigms into a common executable graph and mapping space.
On top of this abstraction, a system performance awareness module jointly estimates the end-to-end latency and energy efficiency of candidate schemes by combining offline stage-level profiling with online network and edge workload sensing.
Guided by these estimates, an SLO-constrained energy-priority scheduler continuously selects the most energy-efficient scheme from the SLO-feasible set under changing runtime conditions.
Finally, with a stage-aware runtime and a lightweight co-inference communication middleware, EcoVLA efficiently executes the selected scheme on real device-edge systems.
The motivating intuition is simple: in robotic control loops, the best collaborative scheme is not the one with the lowest latency or the lowest instantaneous power, but the one that delivers the most output actions per joule while satisfying the SLO budget.
Accordingly, EcoVLA tightly couples unified abstraction, system-aware estimation, and energy-priority scheduling, turning VLA co-inference from a model-specific split problem into a runtime optimization problem.
In this way, the collaborative strategy evolves from a fixed device-edge split point into a continuously adapted stage-level execution plan.
Ultimately, this design enables EcoVLA to maximize Actions/J under real-time constraints across heterogeneous VLA models and deployment conditions.

\section{Unified VLA Co-Inference Design Space}
\label{sec:design-space}

VLA models differ substantially in architecture and execution flow, so implementing a dedicated co-inference mechanism for each model family incurs high engineering cost and yields strategies that hardly transfer across paradigms. To overcome this, EcoVLA organizes heterogeneous VLA models into a unified stage-level co-inference graph, on top of which it defines a unified intermediate-state interface and a co-inference mapping space. This abstraction provides a single system view that performance modeling and adaptive scheduling can directly operate on.

\subsection{Stage-Level Abstraction}
\label{sec:stage-abstraction}

EcoVLA semantically decouples the VLA computation graph and represents a single VLA inference as a directed acyclic graph $G=(V,U)$, where each vertex $v\in V$ is an independently deployable inference stage and each edge $u\in U$ encodes an explicit data dependency between stages. 
Unlike operator-level partitioning, EcoVLA requires every stage to satisfy three system-level conditions: (i)~\textbf{serializable}, i.e., its inputs and outputs can be serialized into a structured intermediate packet; (ii)~\textbf{relocatable}, i.e., the weights and context required for its execution can be pre-loaded or resident on the target device; and (iii)~\textbf{independently executable}, i.e., its invocation on any supported device does not depend on implicit cross-stage state. 
Under this view, both autoregressive VLAs (e.g., visual encoding, policy backbone, and action decoding) and diffusion-based VLAs (e.g., prefix embedding, context prefilling, and action generation) can be represented as attributed stage-level DAGs. 
Integrating a new VLA model only requires mapping its computation graph onto existing stage types and registering the corresponding execution functions, with no need to re-implement co-inference logic.

A stage-level graph alone, however, is not sufficient to support executable device-edge co-inference, because different VLA paradigms maintain inter-stage state in incompatible forms: autoregressive models carry sequential context and KV-cache state, whereas diffusion-based models propagate latent representations together with iterative denoising state. To shield the scheduler and runtime from these paradigm-specific details, EcoVLA further defines a unified intermediate-state interface on every stage edge, which encapsulates the diverse internal states into a structured packet consisting of \textit{stage input}, \textit{stage output}, and the necessary \textit{incremental context}. This packet serves both as the minimal unit of inter-stage transmission and as the system object directly consumed by performance modeling and online scheduling. The design brings two benefits. First, the actual byte size of each packet is explicitly exposed to the performance model, allowing communication cost to be modeled precisely. Second, stage attributes are fully decoupled from scheduling logic: the scheduler only sees stage names, dependencies, and packet sizes, while the runtime dynamically dispatches the corresponding execution function according to stage attributes. As a result, EcoVLA acts as a model-agnostic stage execution framework rather than a pipeline tailored to a specific VLA family.

\subsection{Co-Inference Mapping Space}
\label{sec:mapping-space}

Under the above abstraction, the co-inference design space can be formalized as the set of mapping strategies 
$\Pi=\{\pi \mid \pi:V\rightarrow D\}$ from the stage set $V$ 
to the device set $D$.
This work focuses on the most representative deployment scenario, where multiple heterogeneous robot devices collaborate with a single shared edge server. 
Let $D_{\mathrm{device}}$ denote the set of all robot devices, and let $d_{\mathrm{edge}}$ denote the shared edge server. 
For any inference request $r$ issued by a source device $d\in D_{\mathrm{device}}$, its stages can only be placed on $d$ or $d_{\mathrm{edge}}$.
Therefore, the candidate device set for request $r$ is $D_r=\{d,d_{\mathrm{edge}}\}$, and the raw mapping space is $\Pi_r=\{\pi \mid \pi:V\rightarrow D_r\}$.
For a model with $|V|$ stages, ignoring additional constraints, the size of this raw mapping space is $|\Pi_r|=2^{|V|}$.
In practice, not every stage can be executed on every device in $D_r$. EcoVLA therefore filters $\Pi_r$ by stage--device executability, and defines the feasible mapping space of request $r$ as
\begin{equation}
\Pi_{\mathrm{feas}}^{(r)}=
\left\{
\pi\in\Pi_r
\;\middle|\;
C(v,\pi(v))=1,\;
\forall v\in V
\right\},
\end{equation}
where $C(v,d)$ indicates whether stage $v$ is executable on device $d$, mainly determined by factors such as available memory, operator support, and data-type compatibility. All candidate plans evaluated by the scheduler are drawn from $\Pi_{\mathrm{feas}}^{(r)}$, ensuring that performance modeling and online scheduling are always confined to the deployable subspace.
\section{System Performance Awareness}
\label{sec:perf-model}

Online trial-and-error evaluation of every candidate mapping would incur prohibitive overhead in a heterogeneous device-edge system jointly affected by computation, communication, and queuing. EcoVLA therefore builds a joint performance model that combines offline stage-level profiling, online communication and workload sensing, and post-execution feedback for continuous calibration.

\subsection{Stage-Level Latency and Energy Profiling}
\label{sec:stage-profiling}

For any stage $v$ executed on device $d$, EcoVLA models the predicted execution time as the sum of the steady-state computation time and the stage-boundary system overhead, i.e., $\hat t(v,d)=\hat t^{\mathrm{comp}}(v,d)+\hat t^{\mathrm{sys}}(v,d)$. Here, $\hat t^{\mathrm{comp}}(v,d)$ denotes the steady-state computation time of stage $v$ on device $d$, while $\hat t^{\mathrm{sys}}(v,d)$ captures observable overheads around the stage boundary, including intermediate-packet encoding, receiver-side decoding, runtime dispatch, node queuing, and send-side processing. The corresponding stage-level energy is modeled as $\hat e(v,d)=\hat p(v,d)\cdot \hat t(v,d)$, where $\hat p(v,d)$ is the effective power estimate for executing stage $v$ on device $d$, jointly derived from offline stage-level power profiling and online execution feedback. By jointly modeling latency and energy at the stage--device granularity, EcoVLA can more accurately distinguish the actual cost of different co-inference mappings.

\subsection{Online Communication and Workload Awareness}
\label{sec:online-awareness}

Beyond on-device and edge-side computation, the transmission of intermediate packets across stages placed on different devices is also a critical contributor to inference latency. For a directed edge $\langle v_i,v_j\rangle$, EcoVLA estimates the per-transfer cost from the actual byte size $B_{v_i,v_j}$ of the structured intermediate packet and the current data-plane probing cache as $\hat\tau_{v_i,v_j}=\hat\tau(\pi(v_i),\pi(v_j),B_{v_i,v_j})$.

In addition to communication, edge-side workload also strongly affects collaboration benefits. Under concurrent access from multiple heterogeneous robots, the edge server receives co-inference tasks with diverse stage compositions and deadlines. EcoVLA therefore maintains multi-stage task queues at the edge, and predicts the short-term edge load from request inter-arrival times, SLO constraints, and current queue states. This prediction is used to estimate the queuing delay each stage may experience and to filter out candidate mappings that are infeasible under the current load. In this way, the performance model jointly perceives communication dynamics and edge contention, rather than relying solely on static offline profiles.

\subsection{End-to-End Latency and Energy Estimation}
\label{sec:e2e-estimation}

Building on the above stage-level profiling, communication awareness, and workload awareness, EcoVLA produces a unified end-to-end latency and energy estimate for any candidate mapping $\pi\in\Pi_{\mathrm{feas}}^{(r)}$. For any stage $v$, let $\mathrm{Pa}(v)=\{w\in V\mid \langle w,v\rangle\in U\}$ denote its parent set, $a_{\pi(v)}(t)$ denotes the available time of device $\pi(v)$ at the current moment $t$, and $\mathcal{Q}_{\pi(v),v}(t)$ denotes the queue state associated with stage $v$ on device $\pi(v)$. Together, they characterize the short-term runtime load and serve as inputs to request-level timing estimation and candidate filtering. The predicted start and end times of stage $v$ are then derived recursively as
$
t_v^{\mathrm{start}}
=
\max\!\Big(
a_{\pi(v)}(t),\;
\max_{w\in\mathrm{Pa}(v)}
\big[
t_w^{\mathrm{end}}
+
\mathbb{I}[\pi(w)\neq\pi(v)]\cdot\hat\tau_{w,v}
\big]
\Big),
$
followed by $t_v^{\mathrm{end}}=t_v^{\mathrm{start}}+\hat t(v,\pi(v))$. For the output stage $v_{\mathrm{last}}$, the end-to-end predicted latency of candidate mapping $\pi$ is
\begin{equation}
\hat T(\pi)=t_{v_{\mathrm{last}}}^{\mathrm{end}}+\hat r(\pi), 
\end{equation}
where $\hat r(\pi)$ is a residual term derived from online execution feedback that compensates for systematic biases not fully captured by the offline model. Correspondingly, the end-to-end predicted energy of $\pi$ is
\begin{equation}
\hat E(\pi)
=
\sum_{v\in V}\hat e\big(v,\pi(v)\big)
+
\sum_{\substack{\langle w,v\rangle\in U \\ \pi(w)\neq\pi(v)}}
\hat e^{\mathrm{comm}}_{w,v},
\end{equation}
where $\hat e^{\mathrm{comm}}_{w,v}$ denotes the additional communication energy introduced by transmitting the structured intermediate packet on edge $\langle w,v\rangle$ across devices. Finally, let $L_{\mathrm{act}}(\pi)$ denote the action-chunk length, i.e., the number of effective control steps produced by one inference request under mapping $\pi$. The predicted energy efficiency is then defined as
\begin{equation}
\hat\eta(\pi)=L_{\mathrm{act}}(\pi)/\hat E(\pi). \label{eq:eta}
\end{equation}

In most VLA settings, $L_{\mathrm{act}}(\pi)$ is predetermined by the control window or the action-generation configuration, so the energy-efficiency gap among candidate mappings is dominated by the end-to-end energy cost.
\section{Adaptive Scheduling}
\label{sec:scheduling}

\subsection{Problem Formulation}
\label{sec:problem}

The goal of EcoVLA's runtime scheduler is to select, for each request, the most energy-efficient co-inference plan from the feasible mapping space $\Pi_{\mathrm{feas}}$ while satisfying real-time constraints. Without loss of generality, we express the real-time requirement as a per-request latency budget $T_{\mathrm{SLO}}$, jointly determined by the target control frequency and the action chunk length. Given the end-to-end latency prediction $\hat T(\pi)$ and energy-efficiency prediction $\hat\eta(\pi)$ from Sec.~\ref{sec:perf-model}, the scheduling objective for request $r$ is
\begin{equation}
\pi^{*}(r)=\arg\max_{\pi\in\Pi_{\mathrm{feas}}^{(r)}}\hat\eta(\pi)
\quad
\text{s.t.}
\quad
\hat T(\pi)\le T_{\mathrm{SLO}}.
\end{equation}

When no candidate satisfies the SLO under the current system state, the scheduler degrades to minimizing the violation magnitude:
\begin{equation}
\pi^{\dagger}(r)=\arg\min_{\pi\in\Pi_{\mathrm{feas}}^{(r)}}\hat T(\pi),
\quad
\text{if }\{\pi\mid\hat T(\pi)\le T_{\mathrm{SLO}}\}=\emptyset.
\end{equation}

Since $T_{\mathrm{SLO}}$ is directly determined by task characteristics, users can adjust this threshold to trade off stricter real-time guarantees against higher system energy efficiency.

\subsection{SLO-Constrained Energy-Priority Scheduling}
\label{sec:energy-priority}

For each incoming request $r$, EcoVLA enumerates candidate plans in $\Pi_{\mathrm{feas}}^{(r)}$ that respect device capability, stage residency, and cold-start constraints, and constructs a stage-level execution plan for each candidate. Combining the stage-level latency and energy profiles from Sec.~\ref{sec:perf-model}, the predicted transmission cost of cross-device packets, the earliest available time $a_d(t_0)$ of each node, and the per-stage queue state $\mathcal{Q}_{d,v}(t_0)$, the scheduler recursively estimates the ready, start, and finish times of all stages. For batching-capable edge stages, it further estimates the batch opportunity attainable within the SLO slack from compatible queued requests and per-source request-period predictions, and refines the corresponding stage execution time and per-request energy cost accordingly. As a result, each scheduling decision is not merely a stage mapping $\pi$, but a complete execution plan including stage placement, execution timing, predicted batch sizes, and node occupancy intervals.

After candidate evaluation, EcoVLA performs energy-priority selection under the SLO constraint: it first filters all candidates with $\hat T(\pi)\le T_{\mathrm{SLO}}$ and selects the one with the highest predicted energy efficiency; if no feasible candidate exists, it falls back to the candidate with the smallest predicted latency. Once the chosen plan is fixed, the scheduler registers its node occupancy intervals as reservations, so that subsequent requests can explicitly perceive short-term committed load rather than relying solely on instantaneous queue lengths. After execution, the system uses the observed end-to-end latency, per-stage execution times, boundary system overheads, and communication costs to perform EWMA calibration on the stage-level latency terms, boundary overhead terms, and placement residual in the model. EcoVLA's scheduler is therefore not a static lookup table, but an online decision engine that continuously adapts to link fluctuations, device-resource variations, and edge-load shifts. 

\subsection{Efficient Execution Engine}
\label{sec:execution-engine}

To realize the above stage-level scheduling plan with low additional overhead, EcoVLA does not couple the scheduler with a generic execution backend, but co-designs the execution layer around plan preservability: (1) the runtime must preserve stage boundaries, queue semantics, and batching opportunities, so that the scheduler's decisions based on stage-level timing and load states are not invalidated at execution time; and (2) cross-device transmission must use the structured intermediate packet from Sec.~\ref{sec:design-space} as its basic unit, keeping runtime data units consistent with the communication objects in the performance model. Accordingly, EcoVLA partitions execution support into a stage-aware runtime and a co-inference communication middleware.

\textbf{(1) Schedule-preserving stage runtime.} EcoVLA advances each request in a graph-driven manner across nodes, while within each node a stage-aware runtime maintains stage residency, cold-load policies, and per-stage queues. Once the scheduler determines an execution plan, the runtime triggers stages in dependency order and opportunistically batches requests sharing the same stage and input signature. The goal is not to maximize concurrency, but to keep stage execution cost, queue state, and batch opportunity predictable for the performance model. For batching-capable edge stages, EcoVLA pre-allocates pinned host/device buffers sized for the maximum batch and caches per-signature batch plans to amortize allocation, batch assembly, and host-device staging costs. On top of this, it selectively enables \texttt{torch.compile} and CUDA Graph at the stage level, so that partitioned stages still retain high execution efficiency.

\textbf{(2) Co-inference communication middleware.} Built on the structured packet abstraction defined earlier, EcoVLA introduces a dedicated communication middleware that explicitly separates control and data planes: scheduling control and lightweight metadata travel over a control channel, while inter-stage tensor payloads travel over an independent data channel. Same-node stage edges use local inline passing to avoid extra copies, whereas cross-node edges transmit the tensor tree of each packet via p2p communication built on \texttt{torch.distributed}, combined with pinned memory and pre-allocated buffers to suppress fixed transfer overhead. This avoids the serialization cost of generic object RPC and keeps the runtime data unit strictly aligned with the communication object in the scheduler's model. Send, receive, and downstream-stage readiness checks are organized asynchronously, enabling communication–computation overlap.
\section{Experimental Evaluation}\label{sec:experiment}
\subsection{Experimental Setup}\label{sec:setting}

\textbf{Models, baselines, and request setup.}
We evaluate EcoVLA using five representative VLA models. 
These include the autoregressive models OpenVLA~\cite{kim2024openvla}, as well as the diffusion/denoising-based models $\pi_0$, $\pi_0.5$~\cite{black2025pi05}, SmolVLA~\cite{shukor2025smolvla}, and RDT-1b~\cite{liu2025rdt1b}. 
For diffusion-based VLAs, we set the action chunk size to 20 to prevent a decline in action quality caused by excessively long sequences.
To avoid bias introduced by re-implementation, all single-robot baselines directly use the official inference interfaces and pretrained weights of each model, while EcoVLA performs stage-level device-edge co-inference using the same weights. 
Since our focus is deployment efficiency rather than policy accuracy, we use standardized online requests aligned with each model's official action-generation API, while preserving the original multimodal input formats and tensor shapes, including images, language prompts, robot states, and text embeddings. 
We compare EcoVLA against robot-side local execution (Device-Only), full edge offloading (Edge-Only), Fixed Split co-inference schemes~\cite{jiang2026fast}.

\textbf{Device-edge system and implementation setup.}
We build a real device-edge testbed for VLA co-inference. The edge server is equipped with an NVIDIA RTX 4090, and the robot device uses an NVIDIA Jetson AGX Orin 32GB. EcoVLA uses Ray RPC for the control plane and \texttt{torch.distributed} with \textit{Gloo} for the data plane. Stage weights are preloaded on both nodes to eliminate cold-start interference. 
Power measurements are collected using built-in system profilers on both sides: NVML on the RTX 4090 and \texttt{tegrastats} on the AGX Orin devices. 
Stage-level energy consumption is obtained by integrating the power trace over each inference window. In dynamic-network experiments, we further inject staged link-bandwidth perturbations using Linux \texttt{tc} to evaluate the scheduler's adaptability to changing communication conditions.

\begin{figure}[t]
    \centering
    \includegraphics[width=0.9\linewidth]{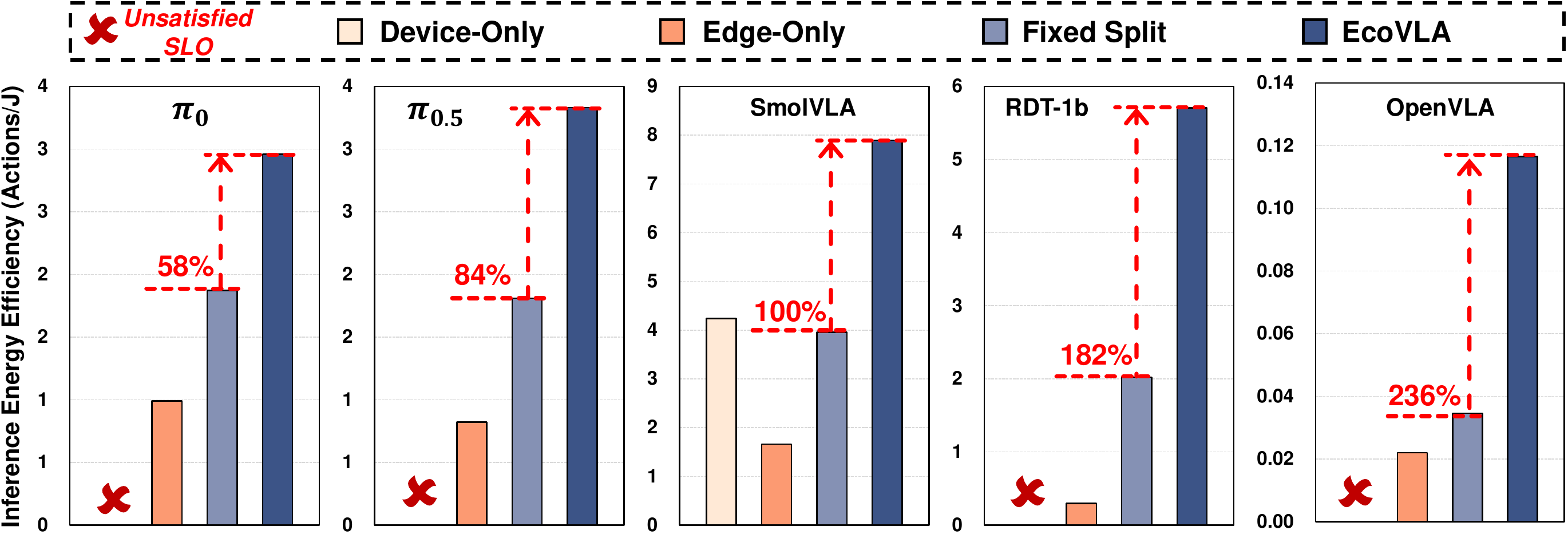}
    \caption{Energy efficiency (Actions/J) comparison across five VLA models on AGX Orin--RTX 4090 under a target action output frequency of 20\,Hz. {\color{red}\ding{55}} indicates the scheme fails to satisfy the real-time constraint.}
    \label{fig:speedups}
    \vspace{0pt}
\end{figure}

\subsection{Experimental Results}\label{sec:results}
Following the real-time control frequency adopted in prior work~\cite{dai2025actionflow}, 
we define the SLO constraint as requiring the action output frequency to reach 20\,Hz. 
This translates to a per-inference latency of  $T_{\text{SLO}} \leq 1$\,s 
for models with an action chunk size of 20.

\textbf{EcoVLA vs. Existing approaches.}
EcoVLA consistently achieves higher energy efficiency among all SLO-satisfying schemes, when compared against feasible baseline—taking the Fixed Split (static partitioning) strategy as a representative baseline shown in Fig.~\ref{fig:speedups} for each model, improving over this baseline by 58\% on $\pi_0$, 84\% on $\pi_{0.5}$, 100\% on SmolVLA, 182\% on RDT-1b, and 236\% on OpenVLA. Notably, although Device-Only can achieve relatively competitive energy efficiency—and even outperforms optimized schemes on certain models—it frequently violates the SLO constraint. Specifically, Device-Only inference fails to meet the SLO on four out of five models, indicating that its high efficiency is achieved at the cost of unacceptable latency. This result highlights a fundamental limitation: on-device hardware alone is insufficient to sustain real-time VLA control, even when its raw energy efficiency appears favorable.
It is worth noting that Fig.~\ref{fig:speedups} reports only the final energy-optimal plan selected by EcoVLA for each model. 
In practice, EcoVLA does not commit to a single Fixed Split, 
but explores a family of SLO-feasible execution plans 
and selects among them according to the optimization priority. 
Taking $\pi_{0.5}$ as a concrete example, 
under the same 1\,s SLO constraint, EcoVLA produces 
a latency-oriented plan (\textit{EcoVLA-fast}: 712\,ms, 8.15\,J, 2.45\,Actions/J) and an energy-oriented plan (\textit{EcoVLA-opt}: $902ms$, $6.0J$, $3.3$ Actions/J). 
Compared with \textit{EcoVLA-fast}, \textit{EcoVLA-opt} reduces energy consumption by $26.3\%$ and improves energy efficiency by $35.8\%$, while both remain within the SLO budget. 
This flexibility stems from EcoVLA's stage-level design space 
and runtime scheduling mechanism, 
which enable continuous adaptation rather than static partitioning.

\begin{figure}[t]
    \centering
    \includegraphics[width=0.75\linewidth]{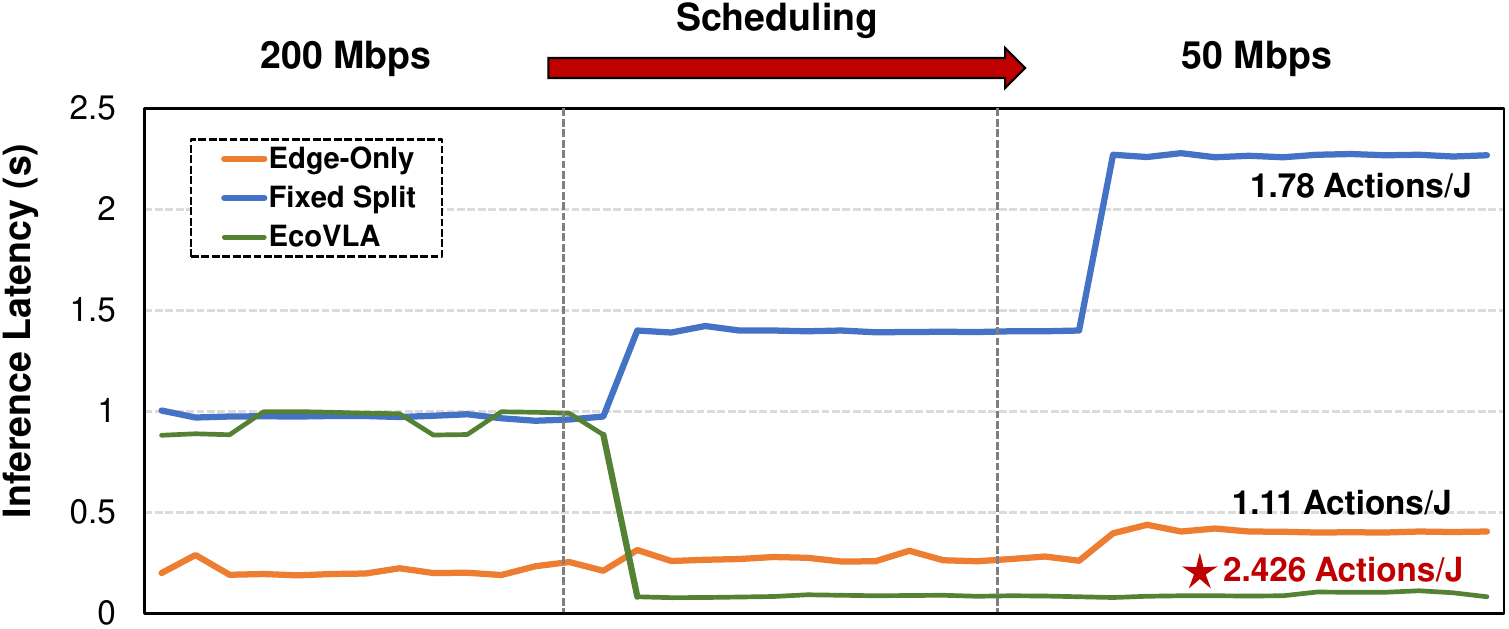}
    \caption{Device-edge co-inference performance variation under network deterioration.}
    \label{fig:timestamp}
    \vspace{0pt}
\end{figure}

\textbf{Evaluation under Network and Edge Workload Variability.}
To evaluate EcoVLA's runtime adaptability, 
we inject a staged bandwidth restriction 
from $200$Mbps to $50$Mbps and trace end-to-end latency over time.
As shown in Fig.~\ref{fig:timestamp}, 
Fixed Split suffers the most severe degradation, 
as its static partition assumes a fixed communication budget 
that no longer holds under reduced bandwidth.
Edge-Only also experiences a moderate latency increase 
due to the higher transmission cost of full-model offloading.
In contrast, EcoVLA detects the bandwidth change online 
and re-schedules to a device-heavier plan, 
stabilizing at a latency of $0.1$ s 
and achieving $2.4$ Actions/J during the degraded phase.
Across light, medium, and heavy load regimes, EcoVLA consistently outperforms both static partitioning and Edge-Only. Compared with static partitioning, it reduces latency/energy by $12.6\%$/$15.8\%$, $22.4\%$/$26.7\%$, and $35.9\%$/$31.8\%$, respectively; compared with Edge-Only, the corresponding gains are $8.9\%$/$10.7\%$, $16.3\%$/$19.5\%$, and $26.8\%$/$24.6\%$. This trend shows that runtime adaptation becomes increasingly valuable as system load grows.

\begin{figure}[t]
    \centering
    \includegraphics[width=1\linewidth]{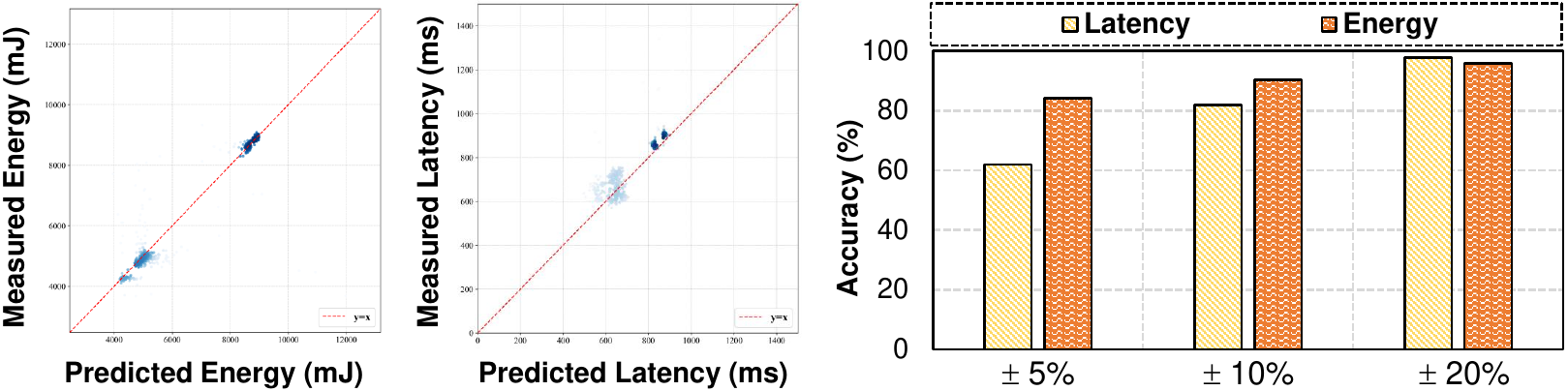}
    \caption{Latency and energy prediction accuracy in co-inference systems.}
    \label{fig:predictor_result}
    \vspace{0pt}
\end{figure}

\textbf{Evaluation of System Performance Awareness.}
We further evaluate the prediction accuracy of EcoVLA's system performance awareness module, which underpins the runtime scheduler's evaluation of candidate co-inference plans. Fig.~\ref{fig:predictor_result} reports the predicted versus measured per-request latency and energy across all candidate mappings explored during scheduling, together with the prediction accuracy under different error tolerance bounds. As shown in the left and middle subfigures, the predicted values closely align with the measured ones along the $y=x$ reference line for both latency and energy, indicating that the joint model faithfully captures stage-level computation, cross-device communication, and queuing effects in the device-edge system. Quantitatively, the latency and energy predictors achieve mean absolute percentage errors (MAPE) of only $6\%$ and $3.8\%$, respectively. The right subfigure further breaks down accuracy by error tolerance: within a $\pm 10\%$ error bound, both latency and energy predictions exceed $80\%$ accuracy, and within a $\pm 20\%$ bound, both surpass $95\%$. Such high prediction fidelity ensures that the scheduler can reliably distinguish SLO-feasible plans from infeasible ones and identify the energy-optimal candidate within the feasible region, thereby providing a solid foundation for the millisecond-level adaptive scheduling decisions.

\section{Conclusion}\label{sec:conclusion}

In this paper, we presented EcoVLA, the first paradigm-agnostic adaptive device-edge co-inference framework for VLA models that targets energy efficiency maximization under real-time constraints. EcoVLA decouples heterogeneous VLA computation graphs into a unified stage-level abstraction, establishing an architecture-agnostic co-inference design space together with a structured intermediate-state interface. On top of this abstraction, EcoVLA builds a joint latency and energy performance model that combines offline stage-level profiling with online communication and edge workload sensing, enabling fast runtime evaluation of candidate co-inference plans. Guided by these estimates, an SLO-constrained energy-priority scheduler continuously selects the most energy-efficient plan from the SLO-feasible region under changing network and load conditions, while a schedule-preserving stage runtime and a dedicated co-inference communication middleware ensure that scheduling decisions are faithfully realized at execution time. Experiments across five representative VLA models on a Jetson AGX Orin--RTX 4090 device-edge platform show that, under a $20$ Hz action output frequency constraint, EcoVLA improves system energy efficiency by up to $236\%$ over fixed co-inference baselines while consistently meeting the SLO, and sustains SLO satisfaction with energy-optimal plans under dynamic network and edge workload variations. In future work, we plan to extend EcoVLA to multi-edge collaborative scenarios and incorporate VLA-specific compression techniques into the stage-level co-inference design space, further pushing the energy-efficiency frontier of real-time embodied AI deployment.

\bibliographystyle{splncs04}
\bibliography{ref}

\end{document}